\documentclass{l4dc2025}

\title[TAB-Fields]{TAB-Fields: A Maximum Entropy Framework for Mission-Aware Adversarial Planning}
\usepackage{caption}
\usepackage{subcaption}
\usepackage{times}
\usepackage{enumitem}
\usepackage{amsmath}
\usepackage[export]{adjustbox} 
\usepackage[utf8]{inputenc}
\usepackage{amssymb}
\usepackage{algorithm2e}
\usepackage{float}
\usepackage{graphicx}
\usepackage{gensymb}
\usepackage{wrapfig}
\usepackage{booktabs}
\usepackage{multirow}
\usepackage{pifont}     
\usepackage{colortbl}
\usepackage{caption}
\usepackage{subcaption}

\RestyleAlgo{semicolon} 
\DontPrintSemicolon     
\SetAlgoNlRelativeSize{-1} 

\SetCommentSty{mycommentfont}

\coltauthor{\Name{Gokul Puthumanaillam\textsuperscript{*}} \Email{gokulp2@illinois.edu}\\
\Name{Jae Hyuk Song\textsuperscript{*}} \Email{jhsong2@illinois.edu}\\
\addr University of Illinois Urbana-Champaign, USA
\AND
\Name{Nurzhan Yesmagambet} \Email{nurzhan.yesmagambet@kaust.edu.sa}\\
\Name{Shinkyu Park} \Email{shinkyu.park@kaust.edu.sa}\\
\addr King Abdullah University of Science and Technology, Saudi Arabia
\AND
\Name{Melkior Ornik} \Email{mornik@illinois.edu}\\
\addr University of Illinois Urbana-Champaign, USA
}

\begin{document}

\maketitle
\begin{abstract}%
Autonomous agents operating in adversarial scenarios face a fundamental challenge: while they may know their adversaries' high-level objectives, such as reaching specific destinations within time constraints, the exact policies these adversaries will employ remain unknown. Traditional approaches address this challenge by treating the adversary's state as a partially observable element, leading to a formulation as a Partially Observable Markov Decision Process (POMDP). However, the induced belief-space dynamics in a POMDP require knowledge of the system's transition dynamics, which, in this case, depend on the adversary's unknown policy. Our key observation is that while an adversary's exact policy is unknown, their behavior is necessarily constrained by their mission objectives and the physical environment, allowing us to characterize the space of possible behaviors without assuming specific policies.
In this paper, we develop Task-Aware Behavior Fields (TAB-Fields), a representation that captures adversary state distributions over time by computing the most unbiased probability distribution consistent with known constraints. We construct TAB-Fields by solving a constrained optimization problem that minimizes additional assumptions about adversary behavior beyond mission and environmental requirements. We integrate TAB-Fields with standard planning algorithms by introducing TAB-conditioned POMCP, an adaptation of Partially Observable Monte Carlo Planning. Through experiments in simulation with underwater robots and hardware implementations with ground robots, we demonstrate that our approach achieves superior performance compared to baselines that either assume specific adversary policies or neglect mission constraints altogether. 

\noindent Evaluation videos and code are available at \href{https://tab-fields.github.io}{https://tab-fields.github.io}. 
\end{abstract}


\section{Introduction}
Effective autonomy in adversarial settings remains a fundamental problem in autonomous systems. \citep{a2, a3, a4}. A core challenge in such settings lies in reasoning about the adversary’s state and its future trajectories, especially when critical aspects of their behavior---such as decision-making policies---are unknown \citep{a1}. This lack of knowledge is further complicated by environmental factors like obstacles, terrain constraints, and dynamic operational constraints.
\begin{figure}[ht!]
    \centering
    \includegraphics[width=0.9\linewidth, clip, trim=0 10 0 120]{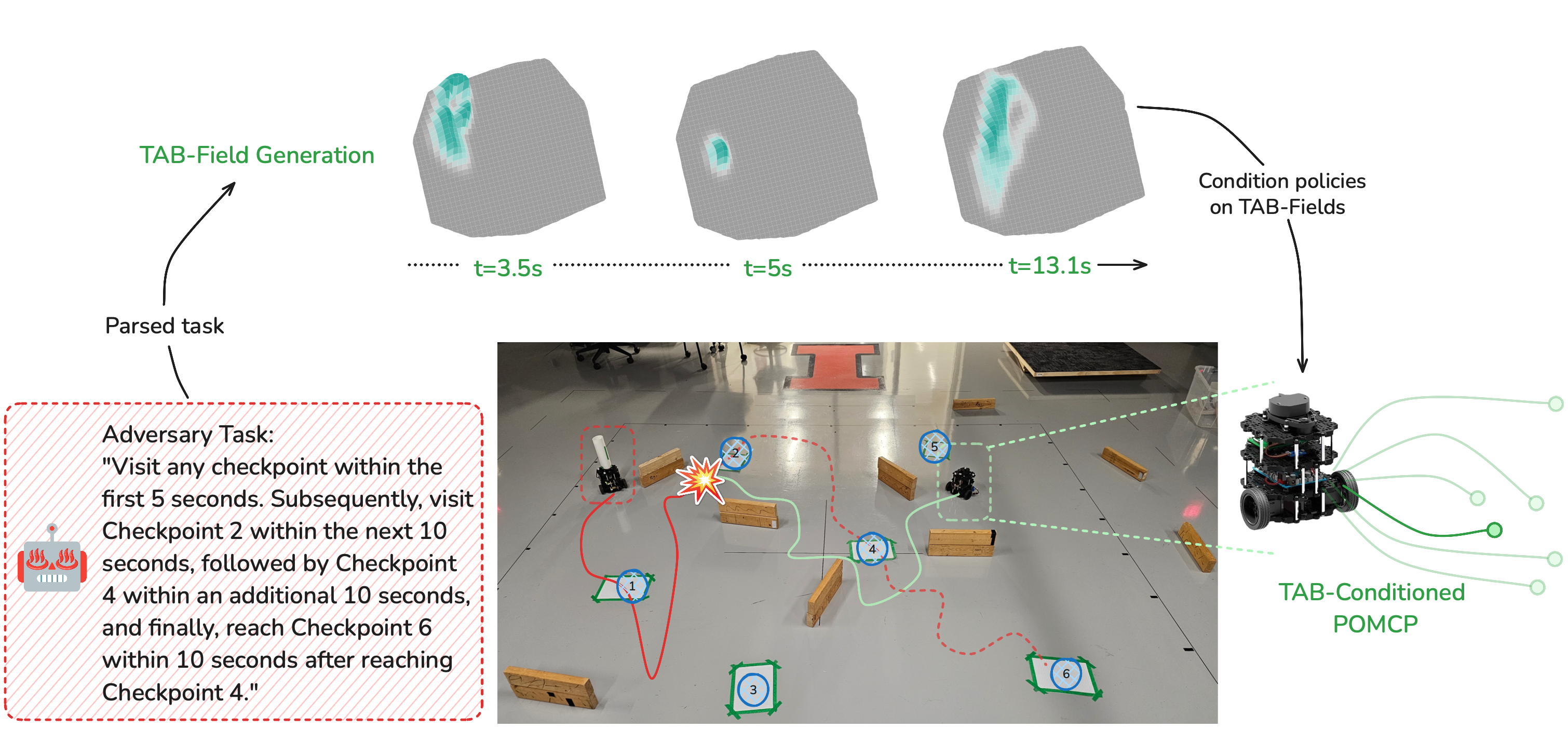}
\caption{Overview of the proposed approach applied to an interception task. The adversary's task is defined by mission objectives and environmental constraints (left). TAB-Fields are generated over time (top) to represent adversary state distributions and integrated into the planning process via TAB-conditioned POMCP (right). The resulting trajectories show the adversary's path (red line), the agent's response (green line), and the interception area (\protect\includegraphics[width=0.03\textwidth, clip, trim=0 100 0 0]{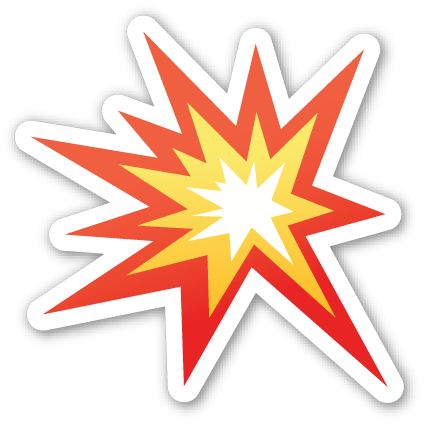}).}
    \label{fig:teaser}
\end{figure}

One way to address this challenge is to treat the adversary’s state as a partially observable element within a broader system \citep{a5, a6}. In this extended state space, the problem can be described as a Partially Observable Markov Decision Process (POMDP) \citep{a7}. POMDPs enable reasoning about uncertainty through belief dynamics---probability distributions over possible states---thereby facilitating structured decision-making. However, a fundamental obstacle arises in this context: the transition dynamics of the system depend on the adversary’s unknown policy, making them inherently indeterminate. Traditional POMDP planning methods rely on a priori knowledge of transition dynamics \citep{a8}, which is unavailable here.

Our key observation is that while an adversary's exact policy is unknown, their behavior is necessarily constrained by their mission objectives and the physical environment. Building on this observation, we propose an alternative approach: instead of assuming a specific adversary policy, we characterize the entire space of possible adversary behaviors that satisfy known mission objectives and environmental constraints. The key idea behind our approach is grounded in the principle of maximum entropy \citep{a9}---\textit{among all probability distributions consistent with the given constraints, the one with the highest entropy offers the most unbiased and comprehensive representation of the current state of knowledge.} Leveraging this principle, we construct a distribution over adversary states that encapsulates the uncertainty in their decision-making while remaining consistent with all available information.

This perspective shifts the focus from predicting specific adversarial behavior to reasoning about them in a mission-aware manner. Through this lens, we introduce Task-Aware Behavior Fields (TAB-Fields), a novel representation that encodes adversary state distributions over time using constrained entropy maximization. As shown in Figure~\ref{fig:teaser}, TAB-Fields capture the evolution of belief states, demonstrating their ability to focus the distribution on regions consistent with mission constraints. TAB-Fields enable us to directly integrate adversary behavior  into the belief update and planning process without relying on explicit policy assumptions or extensive training data.

\noindent \textit{Statement of Contributions.} The primary contribution of this work is TAB-Fields, a novel representation that captures the distribution of possible adversary states through principled entropy maximization subject to mission and environmental constraints. We show how this representation can be effectively integrated with existing planners through TAB-conditioned POMCP, an adaptation that maintains computational tractability while leveraging our structured representation. Through comprehensive evaluation across diverse scenarios in both simulation and hardware experiments, we demonstrate that our approach significantly improves mission-constrained adversarial planning compared to existing methods.

\section{Related Work}
The presented work intersects several research areas in adversarial planning, behavior prediction, and planning under uncertainty. We discuss and highlight how our method differs from prior work.

\noindent\textit{Planning Under Uncertainty.}
Planning in environments with uncertainty has been extensively studied within the framework of Partially Observable Markov Decision Processes (POMDPs) \citep{a7}. Traditional POMDP solvers \citep{c6, c8} rely on known transition and observation models \citep{c9, g5} to perform belief updates and compute optimal policies. However, when the environment includes other agents with unknown policies---such as adversaries---the transition dynamics become partially unknown, complicating standard POMDP approaches \citep{g6, g7}. Several works have extended POMDP frameworks to handle interactions with other agents. Interactive POMDPs \citep{d2, d3} model other agents by maintaining beliefs over their beliefs and policies, but this quickly becomes intractable due to the curse of dimensionality. Decentralized POMDPs \citep{d9} consider multiple cooperative agents, but are less suited for adversarial settings.

\noindent\textit{Modeling Adversary Behavior.} In surveillance and security domains, predicting adversary behavior is critical. Traditional methods \citep{b2, b3, b4} often assume specific models of adversary policies, such as rational decision-makers optimizing a known utility function \citep{h2}. However, these assumptions may not hold in practice, leading to ineffective strategies.
To mitigate this, some approaches use learning-based methods to model adversary behavior from observed data \citep{h4, a4}. While effective when ample data is available, these methods struggle when observations are sparse. Robust planning methods consider worst-case scenarios without relying on specific adversary models \citep{h5, h6}. However, these can be overly conservative.

\noindent\textit{Maximum Entropy Methods for Behavior Prediction.}
The principle of maximum entropy \citep{b6, b9} has been employed to model behavior under uncertainty with known constraints \citep{a9}. In the context of prediction, maximum entropy methods have been used to model motion \citep{g2,g3,g4}, where the goal is to predict likely paths based on environmental features and goal destinations. \citet{b7, b9} applies the idea to design policies for agents under temporal logic constraints by maximizing entropy in constrained MDPs.
Maximum entropy inverse reinforcement learning (IRL) \citep{c1, c2} tackles this problem from a different perspective by recovering reward functions that explain observed behavior, without assuming specific policies. However, IRL requires observed trajectories for learning \citep{g8, g9}, which may not be available in adversarial settings.

\noindent\textit{Belief Planning with Unknown Dynamics}
When transition models are partially unknown, belief planning becomes challenging.
Methods like Robust MDPs \citep{f6, f5} and exploration-exploitation algorithms \citep{f7, f8} address uncertainty by optimizing for the worst-case scenario or learning the dynamics online.
In the context of POMDPs, \citet{f9} propose Monte Carlo POMDPs, where transition probabilities are sampled from a distribution to account for uncertainty. \citet{g1} address model uncertainty by learning models during planning. \citet{f4} extends this work to learn the transition probabilities in dynamic, time-varying POMDPs. 
Our approach avoids the need to learn the adversary's transition dynamics by directly computing the distribution over possible states using the maximum entropy principle and known mission constraints.
Some works consider planning under model uncertainty using robust or risk-sensitive approaches \citep{h1}. However, these methods typically do not leverage known constraints or objectives of other agents.

Our work differs from these approaches by avoiding assumptions about adversary policies or the need for behavior data. Instead of learning from demonstrations like maximum entropy IRL or using nested belief hierarchies as in I-POMDPs, we leverage mission specifications and environmental constraints to compute adversary state distributions through maximum entropy principles. This enables efficient real-time planning without requiring extensive adversary modeling or becoming overly conservative like robust planning methods. By integrating these distributions directly into the POMDP framework, we maintain computational tractability while making informed predictions about adversary behavior based on known constraints.

\section{Preliminaries}\label{sec2}
We consider an ego agent operating in a shared environment with an adversary. The adversary's mission objectives are known, but their exact policy and decision-making processes remain unknown. The environment contains obstacles and operational constraints that affect all agents' feasible actions. Additionally, certain areas provide full observability of the adversary, while in other areas, the adversary is unobservable---a common scenario in surveillance missions where checkpoints or security cameras offer intermittent visibility.

\noindent\textit{Objective.} Our primary problem is to enable the ego agent to plan effectively in this environment without knowledge of the adversary's decision-making process, while maximizing its objectives encoded in the reward function.  Given that the adversary's state is partially observable, we can formulate this as a POMDP. A POMDP typically enables planning through belief space dynamics. However, the transition dynamics of the adversary depend on its unknown policy, making the transition probabilities involving the adversary's state indeterminate---complicating the application of traditional POMDP methods, which typically require known transition models for belief updates and planning. Instead of assuming a specific adversary policy---which could lead to brittle or exploitable behaviors---we seek an approach to reason about the space of possible adversary behaviors.

\noindent\textit{POMDP formulation.} Formally, we define our problem as a POMDP tuple \(\langle \mathcal{S}, \mathcal{A}, \mathcal{O}, T, O, R, \gamma \rangle\). The joint state space \(\mathcal{S}\) encompasses our autonomous agent, adversary, and the environment, with states defined as \(s_t = (s_t^a, s_t^{adv}, s^e)\), where \(s_t^a \in \mathcal{S}^a\) represents our agent's state, \(s_t^{adv} \in \mathcal{S}^{adv}\) represents the adversary's state, and \(s^e \in \mathcal{S}^e\) represents the static environment state.
The action space \(\mathcal{A}\) comprises all available actions for our autonomous agent. The observation space \(\mathcal{O}\) is defined as \(o_t = (o_t^a, o_t^{adv}, o^e)\), where \(o_t^a\) is our agent's fully observable state, \(o_t^{adv}\) represents potentially partial observations of the adversary, and \(o^e\) represents environmental observations. The transition function \(T(s_{t+1} \mid s_t, a_t)\), observation function \(O(o_{t+1} \mid s_{t+1}, a_t)\), reward function \(R(s_t, a_t)\) capturing the agent's objectives [where the reward depends on the both the state of the agent and the adversary], and discount factor \(\gamma\) follow standard POMDP definitions. Note that since the adversary's policy is unknown, the component of $T$ involving $s_{t+1}^{adv}$ is indeterminate.

\noindent\textit{Adversary missions.} This paper's scope considers the adversary's tasks to be specified in natural language defining the mission objectives and environment constraints. These specifications are further processed into an ordered sequence of constraint tuples \(\mathcal{M} = \{(s_{g_i}, t_{c_i}, \text{type}_i, \theta_i)\}_{i=1}^n\), where each tuple specifies a goal state \(s_{g_i}\), temporal constraints \(t_{c_i}\), constraint type (exact time, deadline, until, eventually, or always), and additional constraints \(\theta_i\) such as speed limits or restricted zones. Since many prior works \citep{e1, e6} have focused on this conversion process, we do not explicitly address it here.

The objective is to compute an optimal policy \(\pi^*\) for the ego agent that maximizes the expected cumulative reward \(\mathbb{E}[\sum_{t=0}^{T} \gamma^t R(s_t, a_t) \mid \pi, b_0]\) while maintaining the belief state \(b_t(s)\) over possible states. This belief is updated recursively based on observations through the standard Bayesian update. 
The core challenge lies in performing effective belief updates and planning to maximize the agent's reward function, despite not knowing how the adversary's state evolves over time.

\section{Mission-Aware Adversary Behavior Representation}
To enable effective belief updates and planning, we need a principled way to reason about the adversary's possible states and transitions that is consistent with their known mission objectives and environmental constraints, without assuming knowledge of their specific policies. This problem is closely related to the Schrödinger bridge problem in stochastic processes \citep{e7}, which seeks the most probable evolution of a system between two end-point distributions while minimizing deviation from a reference process \citep{e8}. 

We adopt the principle of maximum entropy \citep{f2}, which states that among all probability distributions satisfying given constraints, the one with the highest entropy is the most unbiased representation of the current state of knowledge. In our context, this means we seek the distribution that satisfies all known mission and environmental constraints while making the minimum number of additional assumptions about the adversary's behavior.

A trajectory of the adversary through the environment can be represented as a sequence of states \(s_{0:T}^{adv} = (s_0^{adv}, \ldots, s_T^{adv})\), where \(s_t^{adv}\) represents the adversary's state at time t. Let \(Q(s_{0:T}^{adv})\) denote a reference probability distribution representing physically feasible transitions based on environmental constraints and dynamics. This reference process, similar to uncontrolled dynamics in KL control \citep{e9}, assigns zero probability to infeasible paths (e.g., through obstacles) and encodes basic motion constraints. 
We seek a distribution \(P(s_{0:T}^{adv})\) that incorporates mission constraints while remaining as close as possible to $Q$, thereby providing a prediction of the adversary's state evolution for use in belief updates. We formulate this as a constrained optimization problem:
\begin{equation}
\begin{aligned}
\min_{P} \quad & D_{\text{KL}}(P \parallel Q) = \sum_{s_{0:T}^{adv}} P(s_{0:T}^{adv}) \log \left(\frac{P(s_{0:T}^{adv})}{Q(s_{0:T}^{adv})}\right) \\
\text{subject to:} \quad & P(s_0^{adv}) = \mu_0(s_0^{adv}), \quad \text{(initial state)} \\
& \mathbb{E}_{P}[f_{\mathcal{M}}(s_{0:T}^{adv})] = c_{\mathcal{M}}, \quad \text{(mission constraints)} \\
& P(s_t^{adv} \in \mathcal{C}) = 0, \quad \forall t \in [0,T], \quad \text{(environment constraints).}
\end{aligned}
\label{eq:kl_minimization}
\end{equation}
\begin{wrapfigure}{r}{0.4\textwidth} 
    \centering
    \includegraphics[width=0.4\textwidth, clip, trim=0 100 0 0]{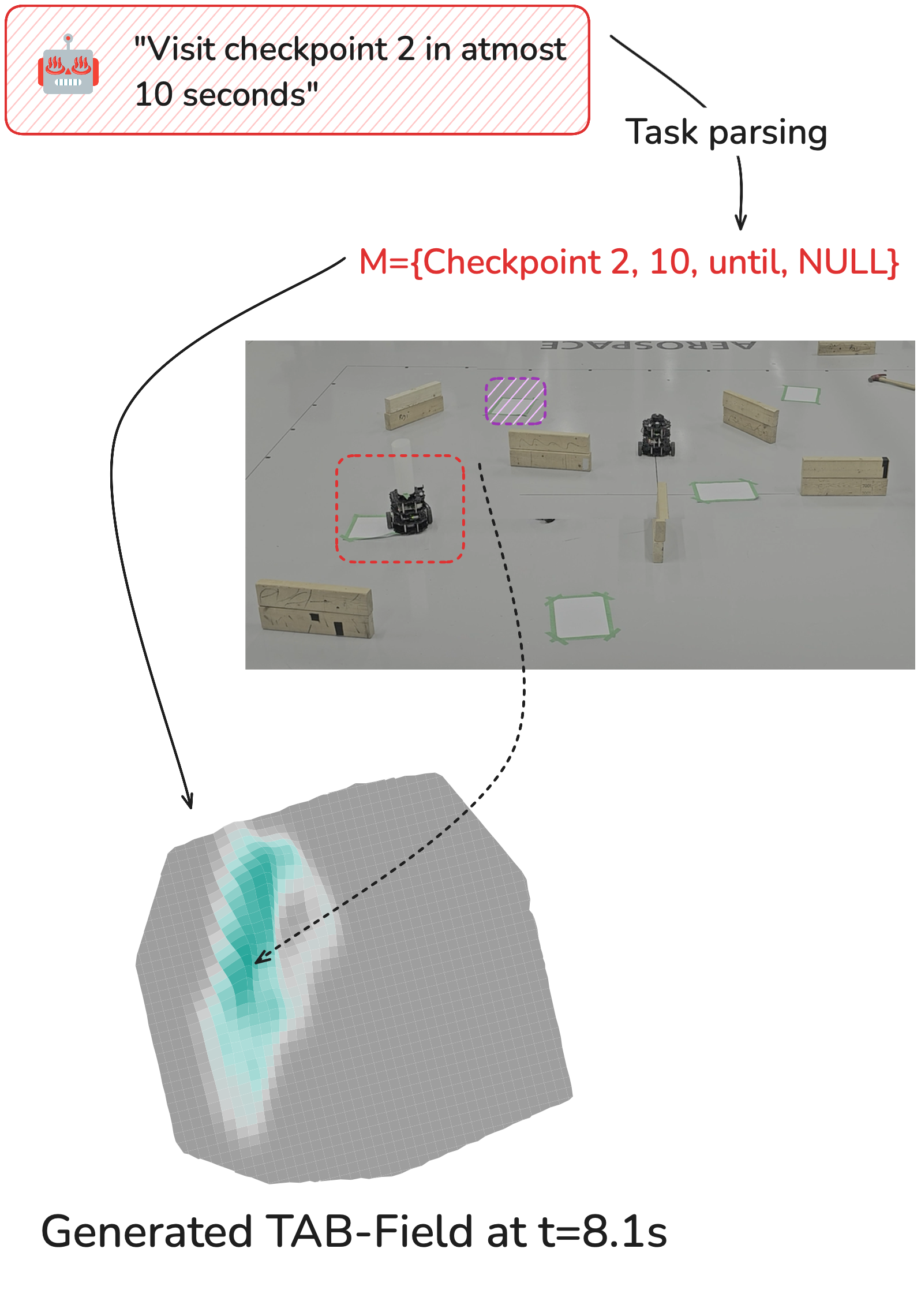}
    \caption{Example mission and its TAB-Field, where darker areas indicate higher probability of adversary presence. Red area denotes adversary start position and purple area indicates the goal checkpoint.}
    \label{fig: }
\end{wrapfigure}
The optimization in equation~(\ref{eq:kl_minimization}) extends  techniques from maximum entropy IRL \citep{c1}, where similar techniques are used to model expert behavior without assuming specific reward functions. In equation~(\ref{eq:kl_minimization}), \(D_{\text{KL}}(P \parallel Q)\) measures how much additional information P contains beyond what is implied by the reference process Q, \(f_{\mathcal{M}}\) represents a vector of constraint functions derived from the mission specification tuples in \(\mathcal{M}\). Each function maps trajectories to binary values indicating whether they satisfy the corresponding requirement. For example, given a mission tuple \((s_g, t_c, \text{exact}, \theta)\), the corresponding constraint function evaluates to 1 if and only if the trajectory reaches state \(s_g\) at time \(t_c\) while satisfying additional requirements \(\theta\). The environment constraints ensure that at each timestep, trajectories through prohibited states $(\mathcal{C})$ have zero probability.

The solution to the optimization problem~(\ref{eq:kl_minimization}) takes a form characteristic of the exponential family of probability distributions \citep{f3}, commonly arising in maximum entropy problems:\[
P^*(s_{0:T}^{adv}) = \frac{1}{Z} Q(s_{0:T}^{adv}) \exp(-\lambda^T f_{\mathcal{M}}(s_{0:T}^{adv}))
\]
where Z is the normalization constant and \(\lambda_i\) are Lagrange multipliers corresponding to each constraint \(f_i\). This solution modifies the reference distribution Q through exponential terms that enforce mission constraints, similar to how the Schrödinger bridge problem modifies a prior process to satisfy endpoint constraints \citep{e8}.

While computing this distribution exactly is intractable due to the high-dimensional state space, we can efficiently compute the marginal distributions \(P^*(s_t^{adv})\) using iterative algorithms from probabilistic graphical models \citep{h7}. These marginal distributions over time form our Task-Aware Behavior Fields (TAB-Fields). 

\subsection{TAB-Conditioned Planning}
Building on TAB-Fields, we now address how to effectively integrate them into the planning process. In our setting, the adversary's transition dynamics depend on their unknown policy, making standard POMDP planning approaches inapplicable. Instead of assuming a specific adversary policy, we use TAB-Fields as a surrogate for the unknown transition dynamics.
The intuitive idea is to perform belief updates using TAB-Fields to predict the adversary's state evolution. Specifically, when a new observation \(o_{t+1}^{adv}\) is received, we update our belief over the adversary's state as:
\[
b_{t+1}(s_{t+1}^{adv}) = \eta \cdot O(o_{t+1}^{adv} \mid s_{t+1}^{adv}) \cdot P(s_{t+1}^{adv})
\]
where \(P(s_{t+1}^{adv})\) is the probability distribution provided by TAB-Fields and \(\eta\) is a normalization constant. When no observations are available, the belief evolves according to TAB-Fields distribution.

\noindent\textit{TAB-POMCP. }While any POMDP solver could potentially be conditioned on TAB-Fields, we demonstrate our approach POMCP \citep{c6} due to its ability to handle large state spaces efficiently and its natural integration with particle-based belief representations. In standard POMCP, particles representing possible states are propagated using known transition dynamics. Our TAB-conditioned variant instead uses TAB-Fields to guide particle propagation---during each simulation step, the next adversary state is sampled from the TAB-Field distribution. This ensures that simulated trajectories remain consistent with mission constraints and environmental limitations.
The action selection process in TAB-POMCP remains unchanged, using UCT to balance exploration and exploitation. However, the value estimates now account for uncertainty in adversary behavior through the TAB-Field distributions rather than assumed transition models. When observations become available, particles are reweighted according to the observation likelihood, but unobserved adversary states continue to evolve according to the TAB-Fields. This approach maintains POMCP's computational efficiency while enabling planning without explicit adversary policy assumptions.

\section{Experiments and Results}
We evaluate our approach through a series of experiments: hardware implementations with ground robots followed by ablation studies in simulation to evaluate performance across larger state spaces.

\noindent\textit{Motivating Scenario.}
Consider a mission where an autonomous ego vehicle must intercept an adversarial agent targeting critical infrastructure. Through intelligence, our agent knows the adversary's task which is defined by mission objectives and environmental constraints. However, the exact policy the adversary will use to execute this mission remains unknown.
 The agent can only observe the adversary's position when it passes through monitored checkpoints, similar to security cameras providing visibility at key locations.

Note that this interception mission for the ego agent represents one instance of our framework. As described in Section~\ref{sec2}, our approach maximizes a reward function capturing the ego agent's objectives. While we focus on interception throughout our experiments as a concrete example, other missions like adversary avoidance or surveillance are equally applicable. 
\subsection{Experimental Setup}
\noindent\textit{Hardware platform.} We implement both the autonomous agent and the adversary using TurtleBot3 Burger platforms, each equipped with an onboard computer and a LDS-01 Lidar. The platforms run ROS2 with a custom navigation package \citep{h8}. Our experimental area includes markers providing precise localization at designated checkpoints. The environment includes obstacles creating restricted zones, while checkpoints are positioned to simulate critical areas which are monitored. 

\noindent\textit{Ego agent and adversary dynamics.}\label{dyn}
The agents operate under differential drive dynamics with state vector $(x, y, \theta)$ representing position coordinates and heading angle. Control inputs are linear velocity $v \in [0, 0.22\text{ m/s}]$ and angular velocity $\omega \in [0, 1.82 \text{ rad/s}]$. A checkpoint-based observation model provides complete adversary state information only at designated locations, simulating security camera coverage at critical points.

\noindent\textit{Adversary missions.} Following the formulation in Section \ref{sec2}, missions are specified in natural language and are encoded into constraint tuples \(\mathcal{M}\) defining goal states, temporal constraints, and additional requirements.

\noindent\textit{Ego agent mission.} The ego agent aims to intercept the adversary before it reaches critical infrastructure. A reward of +50 is given for successful interception within 0.3m, while collisions incur a -30 penalty. A time step penalty of -1 encourages prompt action, and a control penalty of \(-0.1(v^2 + \omega^2)\) discourages abrupt movements.

\subsection{Baselines}
We evaluate TAB-conditioned POMCP against three baseline approaches representing different methods of handling adversary behavior uncertainty. (i) Standard POMCP (S-POMCP) \citep{c6} represents the original algorithm without mission awareness, where adversary transitions are modeled as uniform random movements within physical constraints -- a common baseline that makes no assumptions about adversary behavior. (ii) Fixed-Policy POMCP (FP-POMCP) assumes the adversary follows a deterministic shortest-path policy to mission objectives, representing commonly used simplified models of goal-directed behavior. (iii) MLE-POMCP uses Maximum Likelihood Estimation to derive adversary transition probabilities from mission constraints and observed data, providing a data-driven comparison that attempts to learn adversary behavior patterns.

\subsection{Results and Analysis}
The performance of TAB-POMCP compared to the baseline methods is summarized in Table \ref{tab:comparison1}. TAB-POMCP consistently outperforms all baselines across all adversary mission types.
\begin{table*}[ht!]
\centering
\renewcommand{\arraystretch}{1.2}
\begin{adjustbox}{max width=\textwidth}
\arrayrulecolor{black!30}
\small
\begin{tabular}{p{0.45\textwidth} l@{\hskip 5pt}|@{\hskip 5pt}c c c >{\columncolor{white}}c}
\toprule
\textbf{Adversary Mission Type} & \textbf{Metric} & \textbf{S-POMCP} & \textbf{FP-POMCP} & \textbf{MLE-POMCP} & \cellcolor{teal!20} \textbf{TAB-POMCP} \\\midrule
\textbf{M1: Basic Reachability} & ATCR (\%)~($\downarrow$) & 85.3\% & 78.0\% & 64.7\% & \cellcolor{teal!20}13.3\% \\
\texttt{Reach Checkpoint A within 5s} &  StI (avg)~($\downarrow$) & 1490 & 1103 & 852 & \cellcolor{teal!20}316 \\
\midrule
\textbf{M2: Sequential Objectives} & ATCR (\%)~($\downarrow$) & 91.2\% & 84.7\% & 79.4\% & \cellcolor{teal!20}18.3\% \\
\texttt{Reach Checkpoint A and then Checkpoint B in exactly 5s} &  StI (avg)~($\downarrow$) & 1882 & 1312 & 1013 & \cellcolor{teal!20}380 \\
\midrule
\textbf{M3: Recurrent Objectives} & ATCR (\%)~($\downarrow$) & 88.1\% & 80.6\% & 45.3\% & \cellcolor{teal!20}19.8\% \\
\texttt{Reach Checkpoint A every 5s} &  StI (avg)~($\downarrow$)& 1631 & 1274 & 953 & \cellcolor{teal!20}412 \\
\midrule
\textbf{M4: Restricted Operation Missions} & ATCR (\%)~($\downarrow$) & 82.0\% & 74.6\% & 59.8\% & \cellcolor{teal!20}15.8\% \\
\texttt{Reach Checkpoint A while avoiding the central zone} &  StI (avg)~($\downarrow$) & 1445 & 1102 & 883 & \cellcolor{teal!20}297 \\
\midrule
\textbf{M5: Multi-Objective Missions} & ATCR (\%)~($\downarrow$) & 95.6\% & 88.9\% & 80.2\% & \cellcolor{teal!20}30.9\% \\
\texttt{Mission Combination (Figure 3)} &  StI (avg)~($\downarrow$) & 2312 & 1871 & 1533 & \cellcolor{teal!20}545 \\
\bottomrule
\end{tabular}
\end{adjustbox}
\caption{Performance comparison between TAB-POMCP and baseline methods on Adversary Task Completion Rate (ATCR) and Average Steps to Interception (StI). Results are averaged over 150 experiments per mission type. Example missions are provided below each category.}
\label{tab:comparison1}
\end{table*}

\begin{figure}[htbp]
    \centering
    \begin{minipage}[b]{0.48\textwidth}
        \includegraphics[width=0.9\textwidth, clip, trim=30 50 30 10]{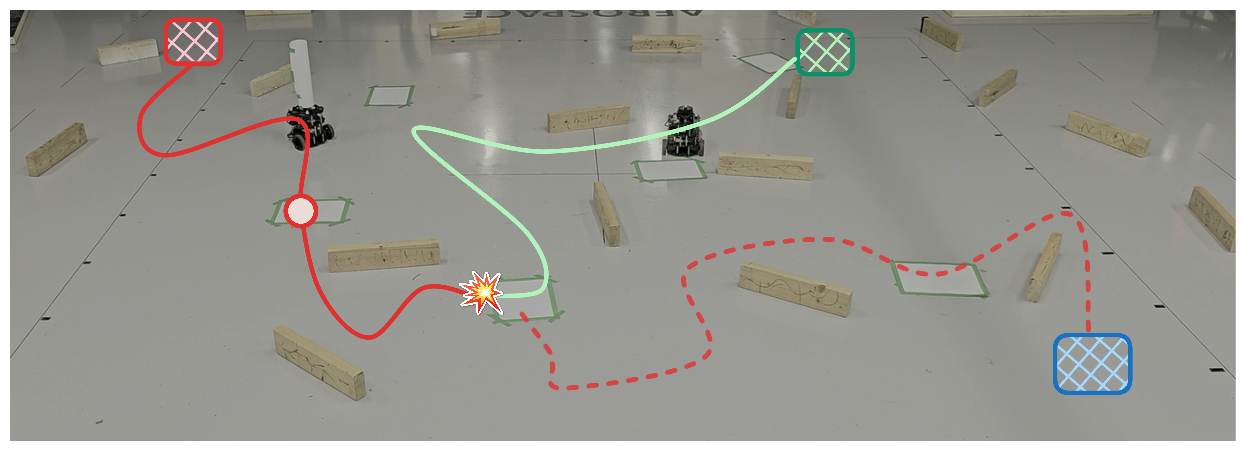}
        \smallskip
        \centerline{(a) TAB-POCMP}
    \end{minipage}
    \begin{minipage}[b]{0.48\textwidth}
        \includegraphics[width=0.9\textwidth, clip, trim=30 50 30 10]{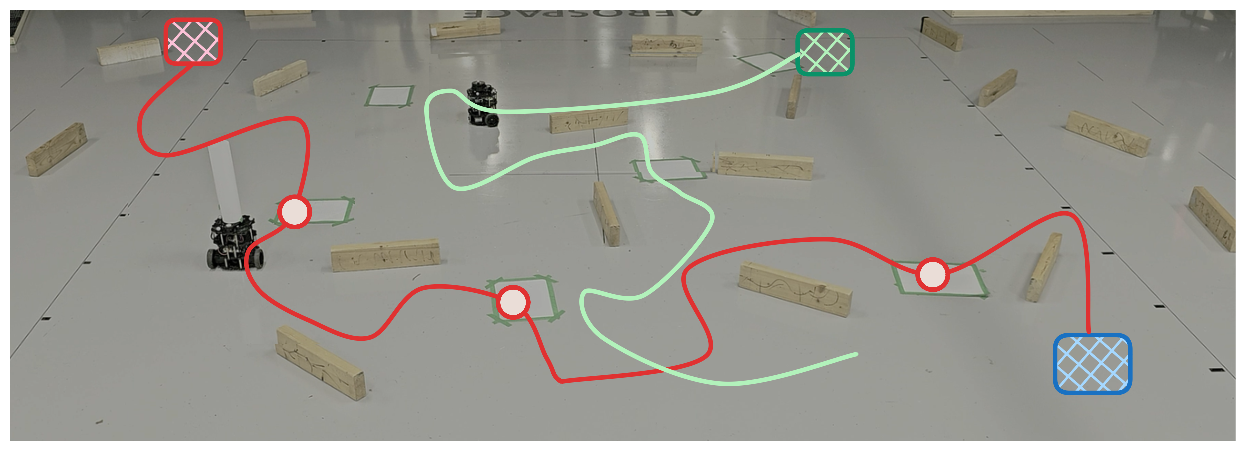}
                \smallskip
        \centerline{(b) S-POCMP}
    \end{minipage}
    
    \begin{minipage}[b]{0.48\textwidth}
        \includegraphics[width=0.9\textwidth, clip, trim=30 50 30 10]{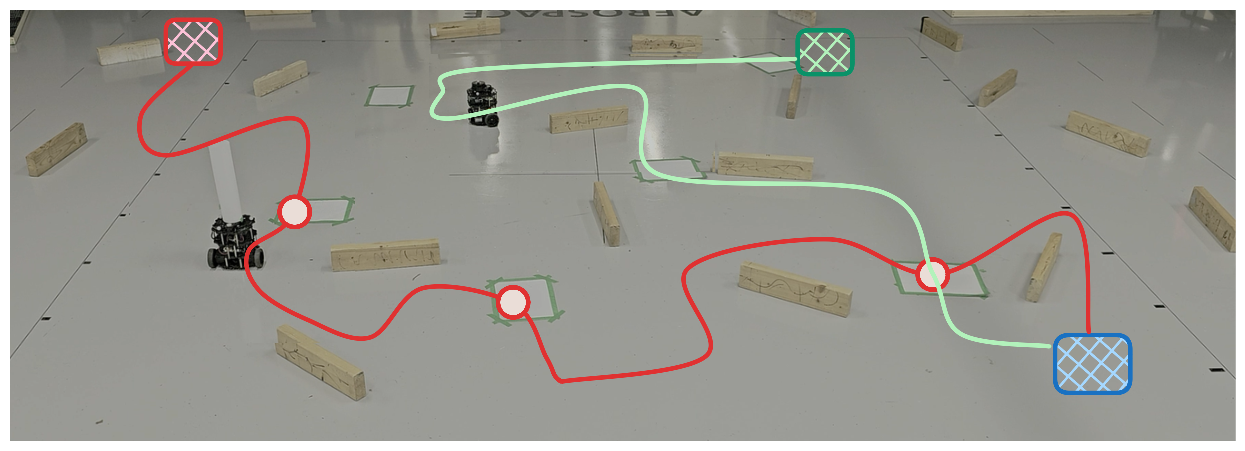}
                \smallskip
        \centerline{(c) FP-POCMP}
    \end{minipage}
    \begin{minipage}[b]{0.48\textwidth}
        \includegraphics[width=0.9\textwidth, clip, trim=30 50 30 10]{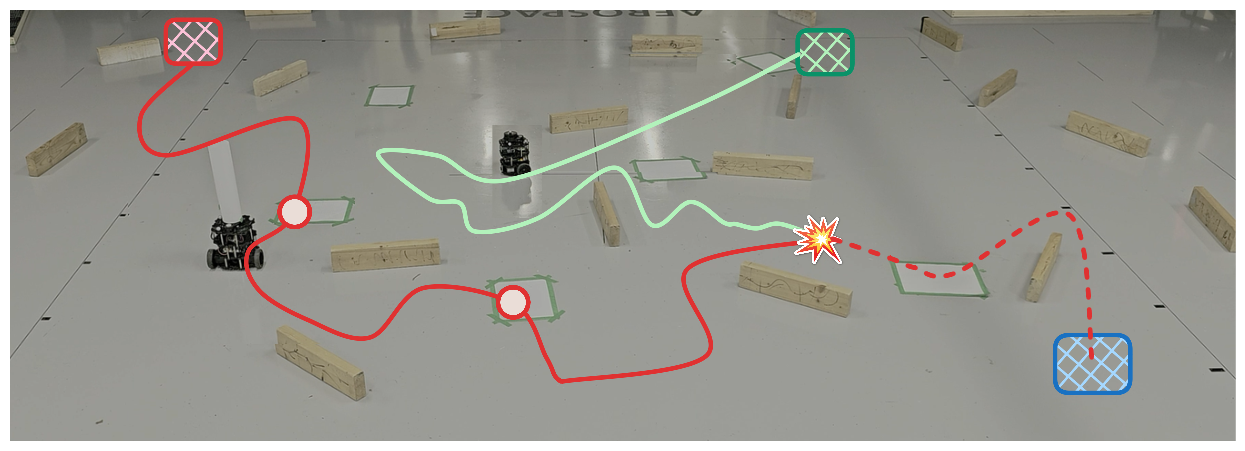}
        \smallskip
        \centerline{(d) MLE-POCMP}
    \end{minipage}
    
    \caption{Comparison of agent (green) and adversary (red) trajectories followed by different approaches. Light red circles indicate full observability points at checkpoints, and \protect\includegraphics[width=0.03\textwidth, clip, trim=0 100 0 0]{pngegg.png} marks the interception area. Adversary mission: Reach target [x,y] after visiting any three different checkpoints, taking no more than 10s between checkpoints, while avoiding the center of the environment.}
    \label{fig:grid}
\end{figure}

\noindent\textit{Impact of conditioning policies on TAB-Fields.}
The comparison between TAB-POMCP and S-POMCP (refer Table~\ref{tab:comparison1}) clearly demonstrates the advantages of incorporating mission constraints into the planning process. S-POMCP, which does not utilize TAB-Fields, exhibits inefficient belief updates, particularly in periods of no observation. This often leads to overly dispersed belief distributions, resulting in ineffective tracking and search patterns. This inefficiency is reflected in consistently higher StI across missions, highlighting the method's inability to effectively narrow down possible adversary states.
In contrast, TAB-POMCP leverages mission constraints to focus belief distributions on regions that align with the adversary’s objectives, even in the absence of observations. This enables more informed and targeted decision-making, leading to significantly higher interception rates. Figure \ref{fig:grid} illustrates this behavior through representative trajectories: while S-POMCP exhibits aimless or overly cautious search patterns, TAB-POMCP efficiently prioritizes high-likelihood regions, demonstrating the impact of mission-aware reasoning.

\noindent\textit{Comparison with alternative mission-aware approaches.}
The results in Table \ref{tab:comparison1} provide key insights into mission-aware planning. As expected, both FP-POMCP and MLE-POMCP outperform S-POMCP, highlighting the value of incorporating mission specifications into the planning process. However, their limitations are evident when examined  closely.
FP-POMCP assumes deterministic, shortest-path behavior for the adversary, which makes it highly brittle in scenarios where the adversary deviates from such paths. This limitation is clearly illustrated in Figure \ref{fig:grid}, where FP-POMCP struggles to adapt to behaviors that does not follow shortest path, leading to significant tracking inefficiencies.
MLE-POMCP, on the other hand, demonstrates better flexibility by learning adversary behavior patterns from data. However, as shown in Figure \ref{fig:grid}, its reliance on sufficient past observations results in poor early-mission performance. The method only improves as it gathers enough data to refine its belief, leaving a critical gap during the initial stages of the mission.
In contrast, TAB-POMCP enables robust performance across all phases of the mission. Unlike FP-POMCP, TAB-POMCP does not assume specific behavior patterns anded by known constraints. Similarly, it avoids MLE-POMCP's reliance on extensive behavioral data, allowing it to excel even in sparse-data scenarios. 

\noindent\textit{Scalability of TAB-conditioned planners.}
We evaluate the scalability of TAB-conditioned planners through high-fidelity underwater vehicle simulations using the BlueROV2 model\footnote{The simulation environment and vehicle dynamics are based on experimental data, available in our open-source repository \href{https://github.com/gokulp01/bluerov2\_gym/}{https://github.com/gokulp01/bluerov2\_gym/}}.
The environment simulates a subsea inspection scenario. The BlueROV2s operate in three-dimensional space with state vector $(x, y, z, \phi, \theta, \psi)$ and corresponding velocities. Similar to the ground robot experiments, the adversary is fully observable only when passing near underwater sensor networks (checkpoint), simulating acoustic or sonar detection zones.
\begin{figure*}[htbp]
\centering
\begin{minipage}{0.48\textwidth}
    \centering
    \renewcommand{\arraystretch}{1.2}
    \begin{adjustbox}{max width=\textwidth}
    \arrayrulecolor{black!30}
    \small
    \begin{tabular}{p{0.20\textwidth}|@{\hskip 5pt}c c c >{\columncolor{white}}c}
    \toprule
    \textbf{Missions} & \textbf{S-POMCP} & \textbf{FP-POMCP} & \textbf{MLE-POMCP} & \cellcolor{teal!20} \textbf{TAB-POMCP} \\\midrule
    \textbf{M1} & 90.1\% & 83.5\% & 70.2\% & \cellcolor{teal!20}19.1\% \\
    \midrule
    \textbf{M2} & 94.8\% & 88.3\% & 82.7\% & \cellcolor{teal!20}24.4\% \\
    \midrule
    \textbf{M3} & 91.7\% & 85.2\% & 50.8\% & \cellcolor{teal!20}27.1\% \\
    \midrule
    \textbf{M4} & 87.5\% & 80.1\% & 65.3\% & \cellcolor{teal!20}21.7\% \\
    \midrule
    \textbf{M5} & 97.2\% & 91.5\% & 85.6\% & \cellcolor{teal!20}43.4\% \\
    \bottomrule
    \end{tabular}
    \end{adjustbox}
    \caption*{Table 2: Performance comparison between different methods on ATCR across different mission categories in an underwater setting. Mission types are the same as that in Table \ref{tab:comparison1} and are abbreviated as M1 through M5.}
    \label{tab:comparison2}
\end{minipage}%
\hfill
\begin{minipage}{0.5\textwidth}
    \centering
    \includegraphics[width=0.7\textwidth, clip, trim=34 50 90 10]{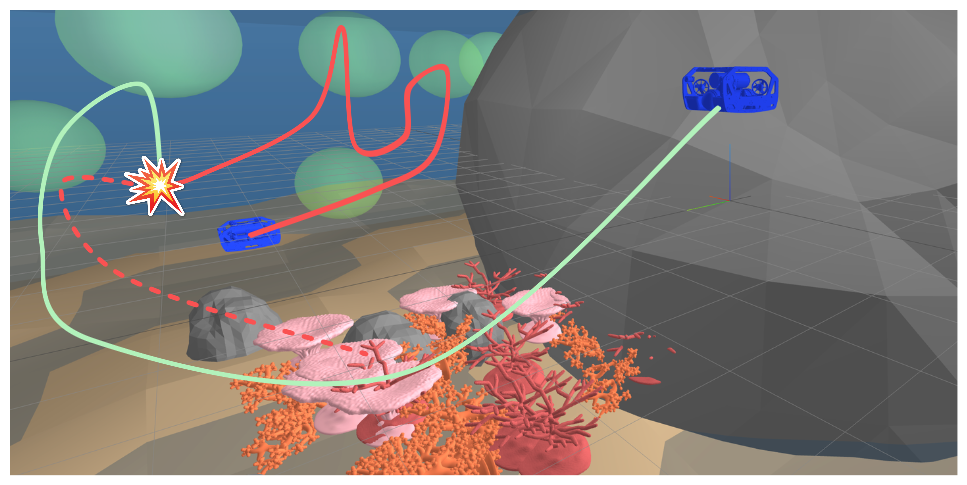}
    \caption{Agent (green) and adversary (red) trajectories using TAB-POMCP. Teal bubbles indicate checkpoints. Adversary task: Reach corals after visiting checkpoints 1, 2, 3 in order, taking no more than 30s between checkpoints.}
    \label{fig:example}
\end{minipage}
\end{figure*}

As shown in Table 2, TAB-POMCP maintains its performance advantage over baselines across all the five mission types (M1-M5). 
The simulation results reveal critical insights about scaling TAB-conditioned policies to higher-dimensional spaces. First, the performance gap between TAB-POMCP and baselines widens as mission complexity increases, particularly in missions with complex temporal dependencies like M5. This suggests that the maximum entropy formulation becomes more valuable precisely when the search space expands. Second, even in the most complex scenarios with multiple interacting constraints (M3), TAB-POMCP maintains a 3-4x improvement in interception efficiency over methods that make explicit policy assumptions.
The key driver behind this scalability is TAB-Fields' ability to automatically identify and exploit mission-constrained regions of the state space. Rather than maintaining beliefs over the full 6-DOF state space, TAB-POMCP effectively ``collapses" the belief to high-probability regions defined by mission constraints. This implicit dimensionality reduction enables efficient planning even as the raw state space grows.

\noindent\textit{Limitations.} Despite the performance benefits, TAB-Field generation incurs additional computational overhead. With efficient parallelized implementation, TAB-POMCP requires approximately 1.4x more computation time compared to standard POMCP. Additionally, while our current formulation handles static obstacles, it does not yet account for dynamic obstacles.

\section{Conclusion}
We presented Task-Aware Behavior Fields (TAB-Fields), a novel approach to reason about adversary behavior in scenarios where mission objectives are known but specific policies remain unknown. Our key contribution lies in recognizing that the maximum entropy principle can characterize the full space of possible adversary behaviors using just mission specifications and environmental constraints, eliminating the need for policy assumptions or hand-crafted rewards.
By solving a constrained optimization problem that minimizes bias beyond known constraints, TAB-Fields provide a distribution over adversary states that captures all feasible behaviors consistent with mission objectives. When integrated with standard planning algorithms through TAB-conditioned POMCP, this representation enables effective decision-making in complex adversarial scenarios. Our experimental results demonstrate significant performance improvements over methods that either make specific policy assumptions or ignore mission constraints. 

\acks{This work of Ornik, Puthumanaillam and Song was supported by the Office of Naval Research under grants N00014-23-1-2651 and N00014-23-1-2505. The work of Park and Yesmagambet was supported by funding from King Abdullah University of Science and Technology (KAUST).}

\bibliography{yourbibfile}

\end{document}